\documentclass{bmvc2k}

\pdfoutput=1

\usepackage{times}
\usepackage{epsfig}
\usepackage{graphicx}
\usepackage{amsmath}
\usepackage{amssymb}

\usepackage{standalone}
\usepackage{adjustbox}
\usepackage{booktabs}
\usepackage[inline]{enumitem}
\usepackage{floatrow}

\usepackage{hyperref}
\hypersetup{%
	pdfauthor={Taylor Mordan, Nicolas Thome, Matthieu Cord, Gilles Henaff},%
	pdftitle={Deformable Part-based Fully Convolutional Network for Object Detection}%
}

\newfloatcommand{capbtabbox}{table}[][\FBwidth]
\newfloatcommand{capbfigbox}{figure}[][\FBwidth]

\title{Deformable Part-based Fully Convolutional Network for Object Detection}

\addauthor{Taylor Mordan}{taylor.mordan@lip6.fr}{1}{2}
\addauthor{Nicolas Thome}{nicolas.thome@cnam.fr}{3}{}
\addauthor{Matthieu Cord}{matthieu.cord@lip6.fr}{1}{}
\addauthor{Gilles Henaff}{gilles.henaff@fr.thalesgroup.com}{2}{}

\addinstitution{
 Sorbonne Universit\'es\\
 UPMC Univ. Paris 06, CNRS, LIP6 UMR 7606\\
 4 Place Jussieu, 75005 Paris, France
}
\addinstitution{
 Thales Optronique S.A.S.\\
 2 Avenue Gay-Lussac, 78990 \'Elancourt, France
}
\addinstitution{
 CEDRIC\\
 Conservatoire National des Arts et M\'etiers\\
 292 Rue St Martin, 75003 Paris, France
}

\runninghead{Mordan, Thome, Cord, Henaff}{Deformable Part-based FCN}

\def\eg{\emph{e.g}\bmvaOneDot}

\def\ie{\emph{i.e}\bmvaOneDot}

\DeclareMathOperator*{\Pool}{Pool}

\begin{document}

\maketitle

\begin{abstract}
	Existing region-based object detectors are limited to regions with fixed box geometry to represent objects, even if those are highly non-rectangular.
	In this paper we introduce DP-FCN, a deep model for object detection which explicitly adapts to shapes of objects with deformable parts. Without additional annotations, it learns to focus on discriminative elements and to align them, and simultaneously brings more invariance for classification and geometric information to refine localization.
	DP-FCN is composed of three main modules: a Fully Convolutional Network to efficiently maintain spatial resolution, a deformable part-based RoI pooling layer to optimize positions of parts and build invariance, and a deformation-aware localization module explicitly exploiting displacements of parts to improve accuracy of bounding box regression.
	We experimentally validate our model and show significant gains. DP-FCN achieves state-of-the-art performances of 83.1\% and 80.9\% on PASCAL VOC 2007 and 2012 with VOC data only.
\end{abstract}

\section{Introduction}
\label{sec:intro}

Recent years have witnessed a great success of Deep Learning with deep Convolutional Networks (ConvNets)~\cite{lecun1989backpropagation, krizhevsky2012imagenet} in several visual tasks. Originally mainly used for image classification~\cite{krizhevsky2012imagenet, simonyan2015very, he2016deep}, they are now widely used for others tasks such as object detection~\cite{girshick2014rich, girshick2015fast, dai2016r, zagoruyko2016a, lin2017feature} or semantic segmentation~\cite{long2015fully, chen2015semantic, li2017fully}. In particular for detection, region-based deep ConvNets~\cite{girshick2014rich, girshick2015fast, dai2016r} are currently the leading methods. They exploit region proposals~\cite{ren2015faster, pinheiro2016learning, gidaris2016attend} as a first step to focus on interesting areas within images, and then classify and finely relocalize these regions at the same time.

Although they yield excellent results, region-based deep ConvNets still present a few issues that need to be solved. Networks are usually initialized with models pre-trained on ImageNet dataset~\cite{russakovsky2015imagenet} and are therefore prone to suffer from mismatches between classification and detection tasks. As an example, pooling layers bring invariance to local transformations and help learning more robust features for classification, but they also reduce the spatial resolution of feature maps and make the network less sensitive to the positions of objects within regions~\cite{dai2016r}, both of which are bad for accurate localization. Furthermore, the use of rectangular bounding boxes limits the representation of objects, in the way that boxes may contain a significant fraction of background, especially for non-rectangular objects.

Before the introduction of Deep Learning into object detection with~\cite{girshick2014rich}, state of the art was led by approaches exploiting Deformable Part-based Models (DPMs)~\cite{felzenszwalb2010object}. These methods are in contrast with region-based deep ConvNets: while the latter relies on strong features learned directly from pixels and exploit region proposals to focus on interesting areas of images, DPM explicitly takes into account geometry of objects by optimizing a graph-based representation and is usually applied in a sliding window fashion over images. Both approaches exploit different hypotheses and seem therefore complementary.

\begin{figure}[t]
	\begin{center}
		\includestandalone[width=\textwidth,mode=image]{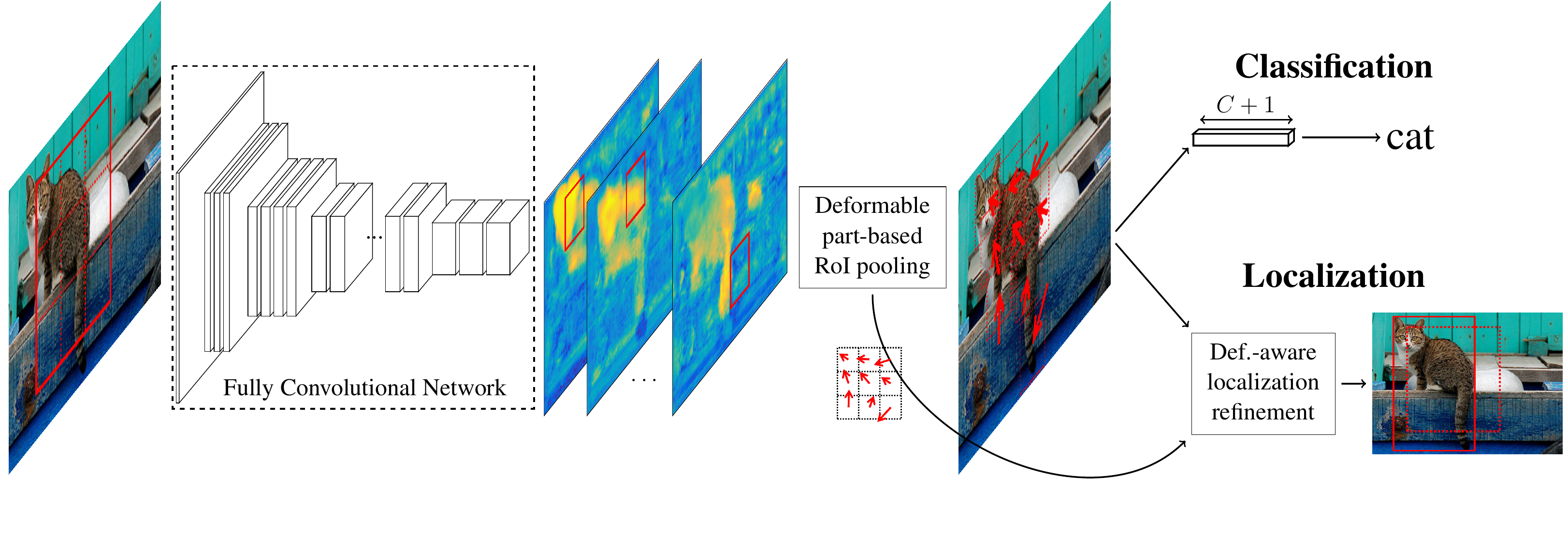}
	\end{center}
	\caption{\textbf{Architecture of DP-FCN.} It is composed of a FCN to extract dense feature maps with high spatial resolution (Section~\ref{sec:fcn}), a deformable part-based RoI pooling layer to compute a representation aligning parts (Section~\ref{sec:dproi}) and two sibling classification and localization prediction branches (Section~\ref{sec:pred}). Initial rectangular region is deformed to focus on discriminative elements of object. Alignment of parts brings invariance for classification and geometric information refining localization \textit{via} a deformation-aware localization module.}
	\label{fig:archi}
\end{figure}

In this paper, we propose Deformable Part-based Fully Convolutional Network (DP-FCN), an end-to-end model integrating ideas from DPM into region-based deep ConvNets for object detection, as an answer to the aforementioned issues. It learns a part-based representation of objects and aligns these parts to enhance both classification and localization. Training is done with box-level supervision only, \ie without part annotation. It improves upon existing object detectors with two key contributions.

The first one is the introduction of a new deformable part-based RoI pooling layer, which explicitly selects discriminative elements of objects around region proposals by simultaneously optimizing latent displacements of all parts (middle of Fig.~\ref{fig:archi}). Using a fixed box geometry must be sub-optimal, especially when objects are not rigid and parts can move relative to each others. Through alignment of parts, deformable part-based RoI pooling increases the limited invariance to local transformations brought by pooling, which is beneficial for classification.

Aligning parts also gives access to their configuration (\ie their positions relative to each others), which brings important geometric information about objects, \eg their shapes, poses or points of view. The second improvement is the design of a deformation-aware localization module (right of Fig.~\ref{fig:archi}), a specific module exploiting configuration information to refine localization. It improves bounding boxes regression by explicitly modeling displacements of parts within the localization branch, in order to tightly fit boxes around objects.

By integrating previous ideas into Fully Convolutional Networks (FCNs)~\cite{he2016deep, dai2016r} (left of Fig.~\ref{fig:archi}), we obtain state-of-the-art results on standard datasets PASCAL VOC 2007 and 2012~\cite{everingham2015the} with VOC data only. We show that those architectures are amenable to an efficient computation of parts and their deformations, and offer natural solutions to keep spatial resolution. The application of deformable part-based approaches is in particular severely dependent on the availability of rather fine feature maps~\cite{savalle2014deformable, girshick2015deformable, wan2015end}.

\section{Related work}
\label{sec:rw}

\paragraph{Region-based object detectors.}

Region-based deep ConvNets are currently the leading approach in object detection. Since the seminal works of R-CNN~\cite{girshick2014rich} and Fast R-CNN~\cite{girshick2015fast}, most of object detectors exploit region proposals or directly learn to generate them~\cite{ren2015faster, gidaris2016attend, pinheiro2016learning}. Compared to sliding window approach, the use of region proposals allows the model to focus computation on interesting areas of images and to balance positive and negative examples to ease learning. Other improvements are now commonly used, \eg using intermediate layers to refine feature maps~\cite{bell2016inside, kong2016hypernet, zagoruyko2016a, lin2017feature} or selecting interesting regions for building mini-batches~\cite{shrivastava2016training, dai2016r}.

\vspace{-1em}
\paragraph{Deformable Part-based Models.}

The core idea of DPM~\cite{felzenszwalb2010object} is to represent each class by a root filter describing whole appearances of objects and a set of part filters to finely model local parts. Each part filter is assigned to an anchor point, defined relative from the root, and move around during detection to model deformations of objects and best fit them. A regularization is further introduced in the form of a deformation cost penalizing large displacements. Each part is then optimizing a trade-off between maximizing detection score and minimizing deformation cost. Final output combines scores from root and all parts. Accurate localization is done with a post-processing step.

Several extensions have been proposed to DPM, \eg using a second hierarchical level of parts to finely describe objects~\cite{zhu2010latent}, sharing part models between classes~\cite{ott2011shared}, learning from strongly supervised annotations (\ie at the part level) to get a better model~\cite{azizpour2012object}, exploiting segmentation clues to improve detection~\cite{fidler2013bottom}.

\vspace{-1em}
\paragraph{Part-based deep ConvNets.}

The first attempts to use deformable parts with deep ConvNets simply exploited deep features learned by an AlexNet~\cite{krizhevsky2012imagenet} to use them with DPMs~\cite{savalle2014deformable, girshick2015deformable, wan2015end}, but without region proposals. However tasks implying spatial predictions (\eg detection, segmentation) require fine feature maps in order to have accurate localization~\cite{lin2017feature}. The fully connected layers were therefore discarded to keep enough spatial resolution, lowering results. We solve this issue by using a FCN, well suited for these kinds of applications as it naturally keeps spatial resolution.
Thanks to several tricks easily integrable into FCNs (\eg dilated convolutions~\cite{chen2015semantic, long2015fully, yu2016multi} or skip pooling~\cite{bell2016inside, kong2016hypernet, zagoruyko2016a}), FCNs have recently been successful in various tasks, \eg image classification~\cite{he2016deep, zagoruyko2016wide, xie2017aggregated}, object detection~\cite{dai2016r}, semantic segmentation~\cite{li2017fully}, weakly supervised learning~\cite{durand2017wildcat}.

\cite{zhang2014part} introduces parts for detection by learning part models and combining them with geometric constraints for scoring. It is learned in a strongly supervised way, \ie with part annotations. Although manually defining parts can be more interpretable, it is likely sub-optimal for detection as they might not correspond to most discriminative elements.

Parts are often used for fine-grained recognition. \cite{lin2015deep} proposes a module for localizing and aligning parts with respect to templates before classifying them, \cite{simon2015neural} finds part proposals from activation maps and learns a graphical model to recognize objects, \cite{zhang2016spda} uses two sub-networks for detection and classification of parts, \cite{sicre2017unsupervised} considers parts as a vocabulary of latent discriminative features decoupled from the task and learns them in an unsupervised way.
Usage of parts is also common in semantic segmentation, \eg~\cite{wang2015joint, dai2016instance, li2017fully}.

The work closest to ours is R-FCN~\cite{dai2016r}, which also uses a FCN to achieve a great efficiency. We improve upon it by learning more flexible representations than with fixed box geometry.
It allows our model to align parts of objects to bring invariance into classification and to exploit geometric information from positions of parts to refine localization.

\section{Deformable Part-based Fully Convolutional Networks}
\label{sec:dpfcn}

In this section, we present Deformable Part-based Fully Convolutional Network (DP-FCN), a deep network for object detection.
It represents regions with several parts that it aligns by explicitly optimizing their positions. This alignment improves both classification and localization: the part-based representations are more invariant to local transformations and the configurations of parts give important information about the geometry of objects.
This idea can be inserted into most of state-of-the-art network architectures. The model is end-to-end trainable without part annotation and adds a small computational overhead only.

The complete architecture is depicted in Fig.~\ref{fig:archi} and is composed of three main modules: \begin{enumerate*}[label=(\roman*)]
\item a Fully Convolutional Network (FCN) applied on whole images,
\item a deformable part-based RoI pooling layer, and
\item two sibling prediction layers for classification and localization\end{enumerate*}.
We now describe all three parts of our model in more details.

\subsection{Fully convolutional feature extractor}
\label{sec:fcn}

Our model relies on a FCN (\eg~\cite{he2016deep, zagoruyko2016wide, xie2017aggregated}) as backbone architecture, as this kind of network enjoys several practical advantages, leading to several successful models, \eg \cite{dai2016r, li2017fully, durand2017wildcat}. First, it allows to share most computation on whole images and to reduce per-RoI layers, as noted in R-FCN~\cite{dai2016r}. Second and most important to our work, it directly provides feature maps linked to the task at hand (\eg detection heatmaps, as illustrated in the middle of Fig.~\ref{fig:archi} and on the left of Fig.~\ref{fig:dproi}) from which final predictions are simply pooled, as done in~\cite{dai2016r,durand2017wildcat}. Within DP-FCN, inferring the positions of parts for a region is done with a particular kind of RoI pooling that we describe in Section~\ref{sec:dproi}.

The fully convolutional structure is therefore suitable for computing responses of all parts for all classes as a single map for each of them. A corresponding structure is used for localization. The complete representation for a whole image (classification and localization maps for each part of each class) is obtained with a single forward pass and is shared between all regions of the same image, which is very efficient.

Since relocalization of parts is done within feature maps, the resolution of those maps is of practical importance. FCNs contain only spatial layers and are therefore well suited for preserving spatial resolution, as opposed to networks ending with fully connected layers, \eg~\cite{krizhevsky2012imagenet, simonyan2015very}. Specifically, if the stride is too large, deformations of parts might be too coarse to describe objects correctly.
We reduce it by using dilated convolutions~\cite{chen2015semantic, long2015fully, yu2016multi} on the last convolution block and skip pooling~\cite{bell2016inside, kong2016hypernet, zagoruyko2016a} to combine the last three.

\subsection{Deformable part-based RoI pooling}
\label{sec:dproi}

\begin{figure}[t]
	\begin{center}
		\begin{tabular}{@{}lr@{}}
			\parbox[c][1em][c]{.73\textwidth}{\includestandalone[width=.73\textwidth,mode=image]{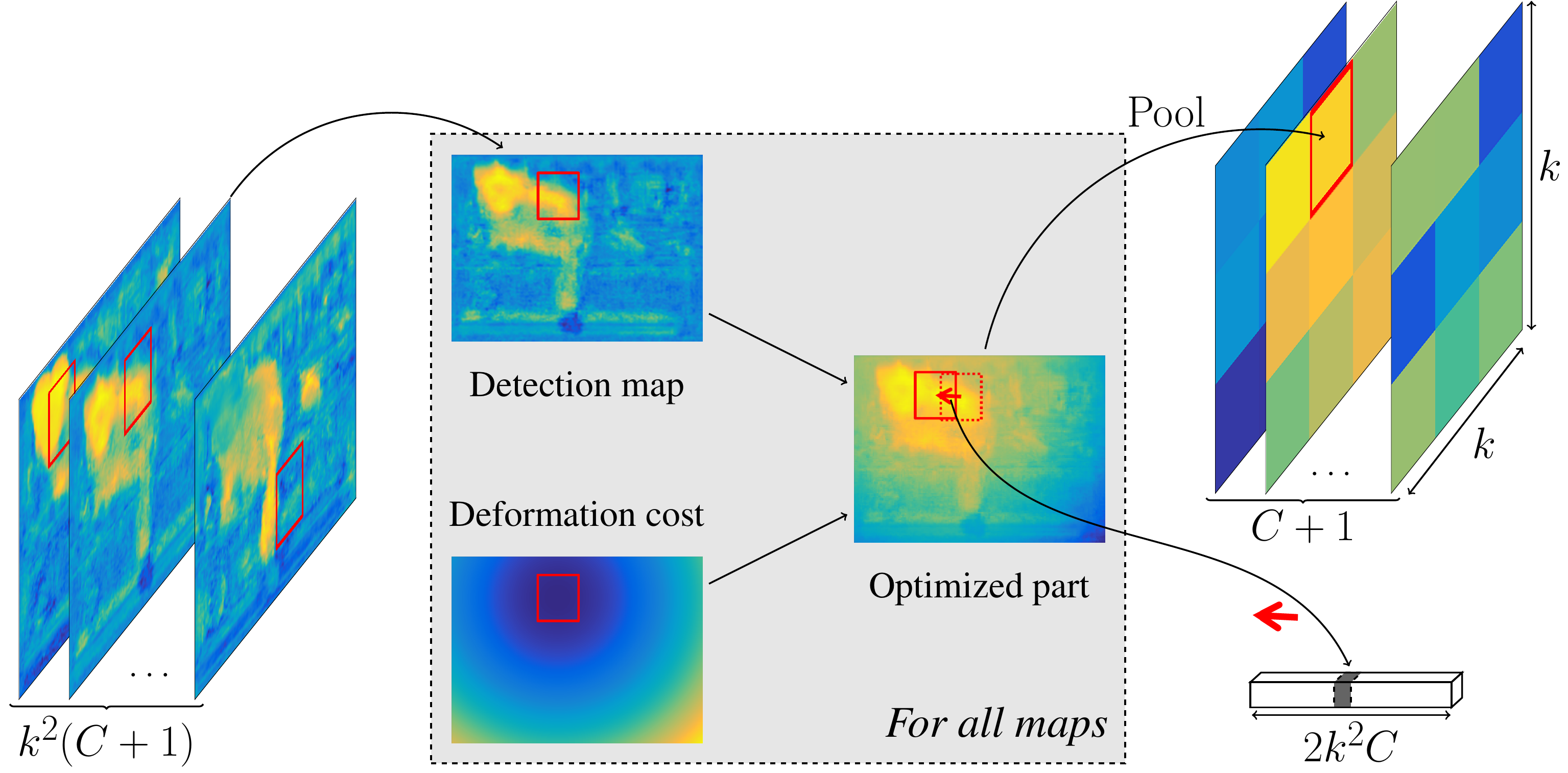}} & \begin{tabular}{@{}c@{}}\includegraphics[width=.235\textwidth]{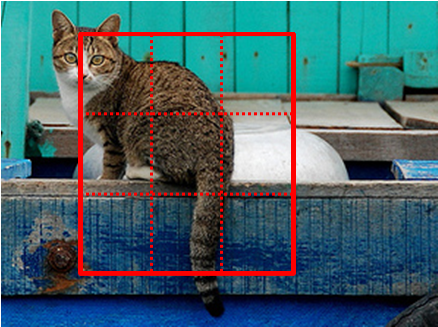}\\\includegraphics[width=.235\textwidth]{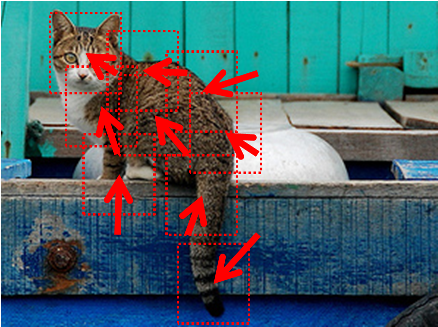}\end{tabular}\\
		\end{tabular}
	\end{center}
	\caption{\textbf{Deformable part-based RoI pooling (left).} Each input feature map corresponds to a part of a class (or background). Positions of parts are optimized separately within detection maps with deformation costs as regularization, and values are pooled within parts at the new locations. Output includes a map for each class and the computed displacements of parts, to be used for localization. \textbf{Illustration of deformations (right).} Parts are moved from their initial positions to adapt to the shape of the object and better describe it.}
	\label{fig:dproi}
\end{figure}

The aim of this layer is to divide region proposals $R$ into several parts and to locally relocalize these to best match shapes of objects (see Fig.~\ref{fig:dproi}).
Each part then models a discriminative local element and is to be aligned at the corresponding location within the image.
This deformable part-based representation is more invariant to transformations of objects because the parts are positioned accordingly and their local appearances are stable~\cite{felzenszwalb2010object}. This is especially useful for non-rigid objects, where a box-based representation must be sub-optimal.

The separation into parts is done with a regular grid of size $k \times k$ fitted to regions~\cite{girshick2015fast,dai2016r}. Each cell $(i,j)$ is then interpreted as a distinct part $R_{i,j}$. This strategy is simple yet effective~\cite{zhu2010latent, wan2015end}. Since the number of parts (\ie $k^2$) is fixed as a hyper-parameter, it is easy to have a complete detection heatmap $z_{i,j,c}$ already computed for each region $(i,j)$ of each class $c$ (left of Fig.~\ref{fig:dproi}).
Parts then only need to be optimized within corresponding maps.

The deformation of parts draws ideas from the original DPM~\cite{felzenszwalb2010object}: it allows parts to slightly move around their reference positions (partitions of the initial regions), selects the optimal latent displacements, and pools values from selected locations.
The pooled score $p^R_c(i,j)$ for part $(i,j)$ and class $c$ is a trade-off between maximizing the score on the feature map and minimizing the displacement $(dx,dy)$ from the reference position (see Fig.~\ref{fig:dproi}):
\begin{equation}
	p^R_c(i,j) = \max_{dx, dy} \left[ \Pool_{(x,y) \in R_{i,j}} z_{i,j,c}(x+dx,y+dy) - \lambda^{def} \left( dx^{2} + dy^{2} \right) \right]
	\label{eq:dproi}
\end{equation}
where $\lambda^{def}$ represents the strength of the regularization (small deformations), and $\Pool$ is an average pooling as in~\cite{dai2016r}, but any pooling function could be used instead. The deformation cost is here the squared distance of the displacement on the feature map, but other functions could be used equally.
Implementation details can be found in Appendix~\ref{sec:impl_dproi}.

During training, deformations are optimized without part-level annotations. Displacement computed during the forward pass are stored and used to backpropagate gradients at the same locations.
We further note that the deformations are computed for all parts and classes independently.
However, no deformation is computed for the \textit{background} class: they would not bring any relevant information as there is no discriminative element for this class.
The same displacements of parts are used to pool values from the localization maps.

$\lambda^{def}$ is directly linked to the magnitudes of the displacements of parts, and therefore to the deformations of RoIs too, by controlling the squared distance regularization (\ie preference for small deformations). Increasing it puts a higher weight on the regularization and effectively reduces displacements of parts, but setting it too high prevents parts from moving and removes the benefits of our approach. It is noticeable this deformable part-based RoI pooling is a generalization of position-sensitive RoI pooling from~\cite{dai2016r}. Setting $\lambda^{def} = +\infty$ clamps $dx$ and $dy$ to 0, leading to the formulation of position-sensitive RoI pooling:
\begin{equation}
	p^R_c(i,j) = \Pool_{(x,y) \in R_{i,j}} z_{i,j,c}(x,y).
	\label{eq:psroi}
\end{equation}
On the other hand, setting $\lambda^{def} = 0$ removes regularization and parts are then free to move. With $\lambda^{def}$ too low, the results decrease, indicating that regularization is practically important. However the results appeared to be stable within a large range of values of $\lambda^{def}$.

\subsection{Classification and localization predictions with deformable parts}
\label{sec:pred}

\begin{figure}[t]
	\begin{center}
		\includestandalone[width=.75\textwidth,mode=image]{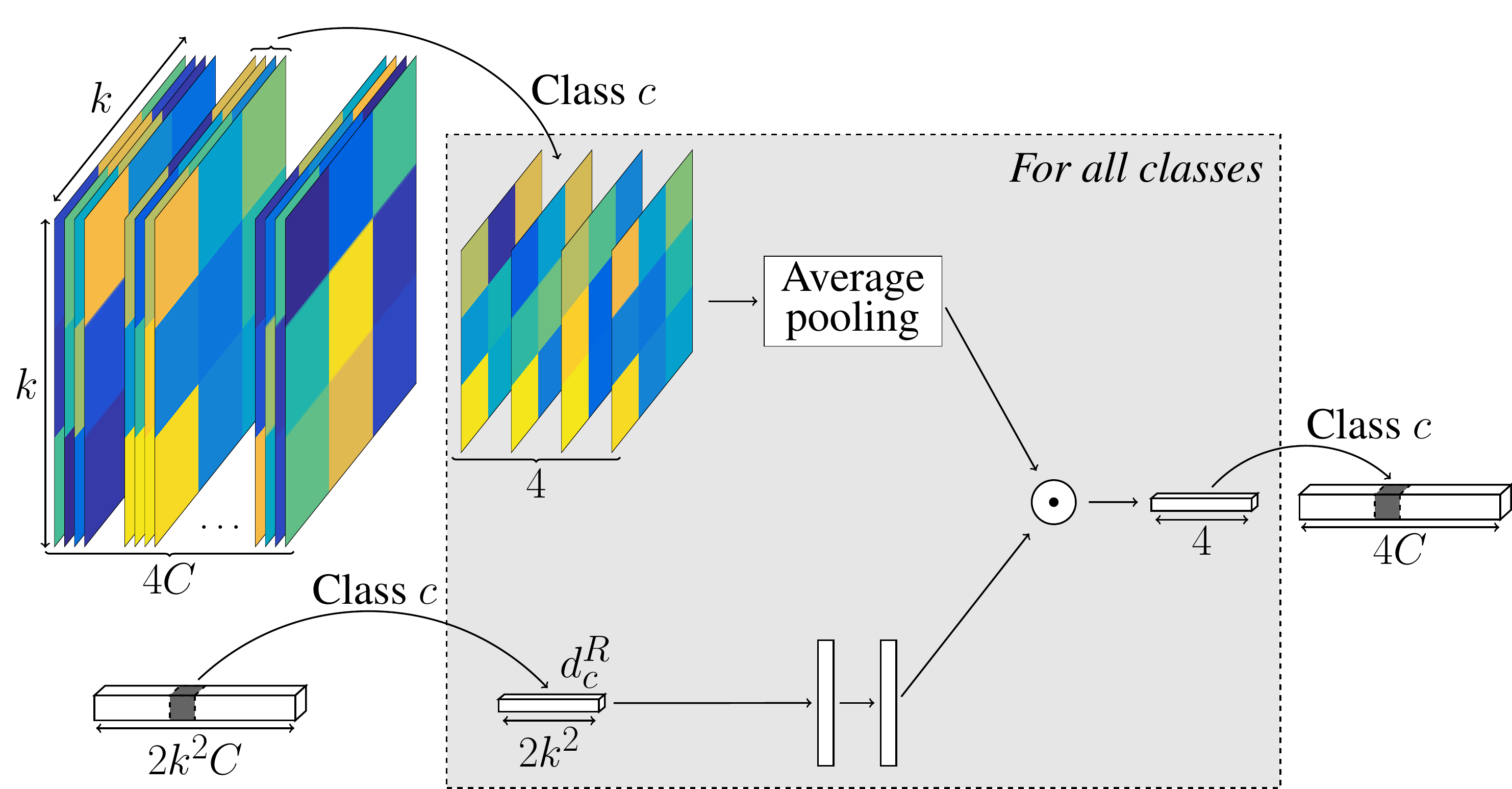}
	\end{center}
	\caption{\textbf{Deformation-aware localization refinement.} Relocalizations of bounding boxes obtained by averaging pooled values from localization maps (upper path) do not benefit from deformable parts. To do so, displacements of parts are forwarded through two fully connected layers (lower path) and are element-wise multiplied with previous output to refine it, separately for each class. Localization is done with 4 values per class, following~\cite{girshick2014rich, girshick2015fast}.}
	\label{fig:loc_ref}
\end{figure}

Predictions are performed with two sibling branches for classification and relocalization of region proposals as is common practice~\cite{girshick2015fast}. The classification branch is simply composed of an average pooling followed by a SoftMax layer. This is the strategy employed in R-FCN~\cite{dai2016r}, however the deformations introduced before (with deformable part-based RoI pooling) bring more invariance to transformations of objects and boost classification.

Regarding localization, we also use an average pooling to compute a first localization output from corresponding features. However, the configuration of parts (\ie their positions relative to each others) is obtained as a by-product of the alignment of parts performed before. It gives rich geometric information about the appearances of objects, \eg their shapes or poses, that can be used to enhance localization accuracy.

To that end we introduce a new deformation-aware localization refinement module (see Fig.~\ref{fig:loc_ref}). For each region $R$, we extract the feature vector $d_c^{R}$ of displacements $(dx,dy)$ for all parts of class $c$ (as shown on Fig.~\ref{fig:dproi}) and use it to refine previous output for the same class. $d_c^{R}$ is forwarded through two fully connected layers and is then element-wise multiplied with the first values to yield the final localization output for this class.
Since refinement is mainly geometric, it is done for all classes separately and parameters are shared between classes.

\section{Experiments}

\subsection{Ablation study}
\label{sec:ablation}

\paragraph{Experimental setup.}

We perform this analysis with the fully convolutional backbone architecture ResNet-50~\cite{he2016deep} and exploit the region proposals computed by AttractioNet~\cite{gidaris2016locnet, gidaris2016attend} released by the authors. We use $k \times k = 7 \times 7$ parts, as advised by the authors of R-FCN~\cite{dai2016r}.
Setting of all others hyper-parameters can be found in Appendix~\ref{sec:xpsetup_ab}.

All experiments in this section are conducted on the PASCAL VOC 07+12 dataset~\cite{everingham2015the}: training is done on the union of the 2007 and 2012 trainval sets and testing on the 2007 test set. In addition to the standard mAP@0.5 (\ie PASCAL VOC style) metric, results are also reported with the mAP@0.75 and mAP@[0.5:0.05:0.95] (\ie MS COCO style) metrics to thoroughly evaluate the effects of proposed improvements.

\vspace{-1em}
\paragraph{Comparison with R-FCN.}

\begin{table}[b]
	\begin{center}
		\begin{tabular}{@{}lcc|ccc@{}}
			\toprule
			Model & Deformations & \begin{tabular}{@{}c@{}}Localization\\refinement\end{tabular} & \begin{tabular}{@{}c@{}}mAP@\\0.5\end{tabular} & \begin{tabular}{@{}c@{}}mAP@\\0.75\end{tabular} & \begin{tabular}{@{}c@{}}mAP@\\{[}0.5:0.95]\end{tabular} \\
			\midrule
			R-FCN & & & 73.7 & 38.3 & 39.8 \\
			 & \checkmark & & 75.8 & 38.8 & 40.4 \\
			DP-FCN & \checkmark & \checkmark & \textbf{76.1} & \textbf{40.9} & \textbf{41.3} \\
			\bottomrule
		\end{tabular}
	\end{center}
	\caption{\textbf{Ablation study of DP-FCN} on PASCAL VOC 2007 test in average precision (\%). Without deformable part-based RoI pooling nor localization refinement module, it is equivalent to R-FCN (the reported results are those of our implementation with the given setup).}
	\label{tab:rfcn}
\end{table}

\begin{figure}[t]
	\centering
	\null
	\hfill
	\includegraphics[width=.379\textwidth]{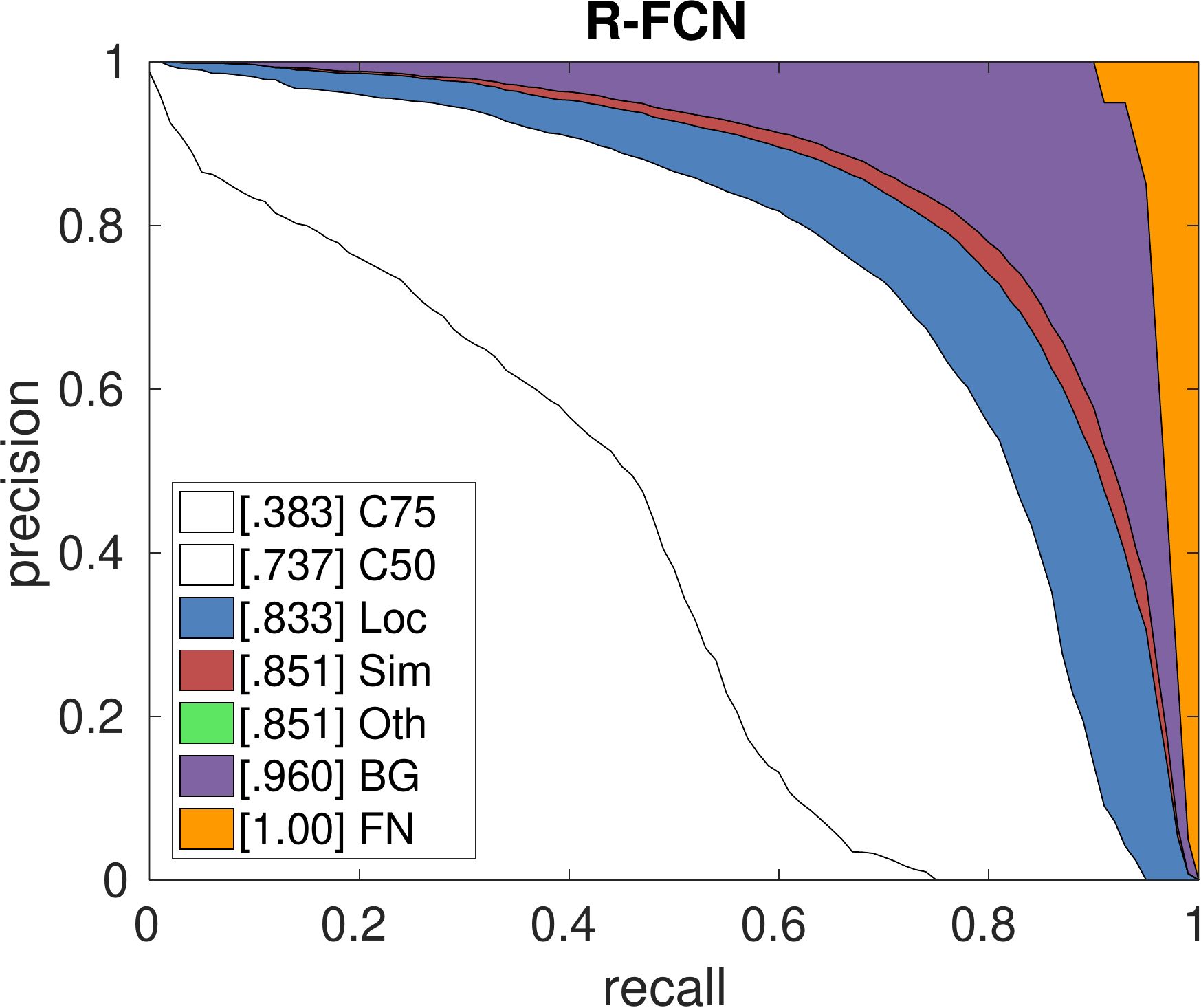}
	\hfill
	\includegraphics[width=.379\textwidth]{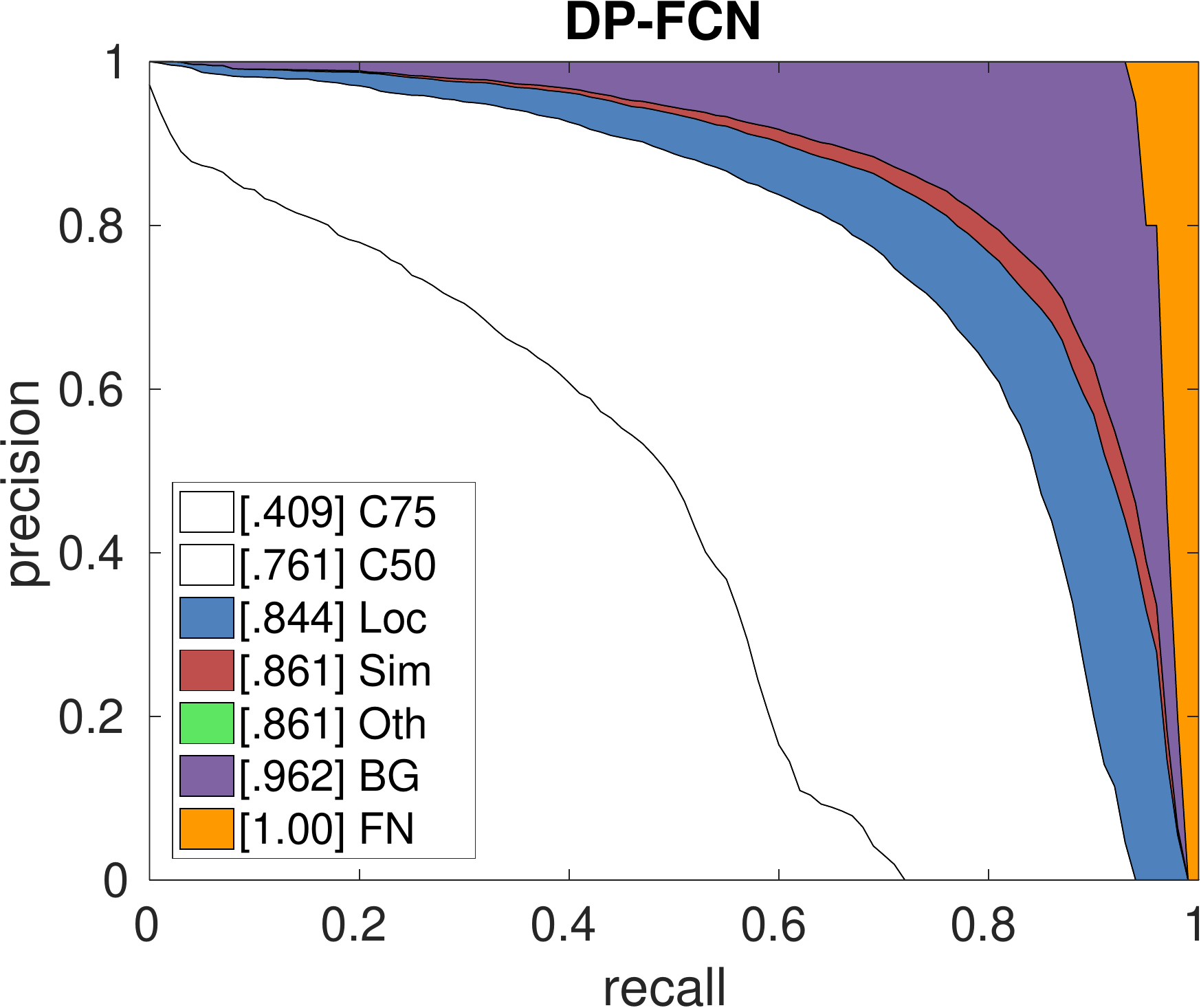}
	\hfill
	\null
	\caption{\textbf{Precision-recall curves for R-FCN (left) and DP-FCN (right).} Detailed analysis of false positives on unseen VOC07 test images averaged over all categories.}
	\label{fig:pr}
\end{figure}

Performances of our implementation of R-FCN~\cite{dai2016r} with the given setup are shown in the first row of Tab.~\ref{tab:rfcn}. Adding the deformable part-based RoI pooling to R-FCN (second row of Tab.~\ref{tab:rfcn}) improves mAP@0.5 by 2.1 points.
Indeed, this metric is rather permissive so the localization does not need to be very accurate: we see that the gain on mAP@0.75 is much smaller. The improvements are therefore mainly due to a better recognition, thus validating the role of deformable parts.
With the localization refinement module (third row of Tab.~\ref{tab:rfcn}), the mAP@0.5 has only a small improvement, because localization accuracy is not a issue. However, it further improves mAP@0.75 by 2.1 points (\ie 2.6 points with respects to R-FCN), validating the need for such a module. This confirms that aligning parts brings geometric information useful for localization.

\begin{figure}[H]
	\centering
	\includegraphics[height=3.3cm]{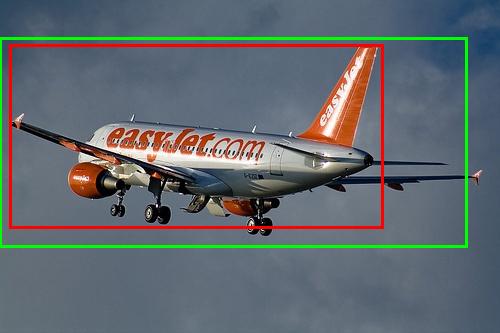}
	\hfill
	\includegraphics[height=3.3cm]{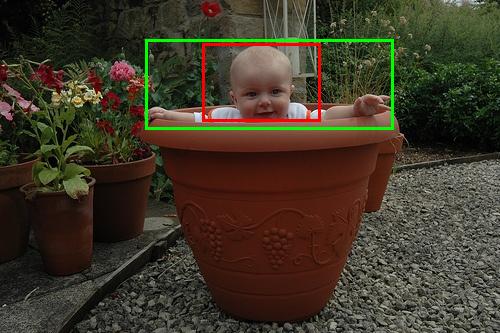}
	\hfill
	\includegraphics[height=3.3cm]{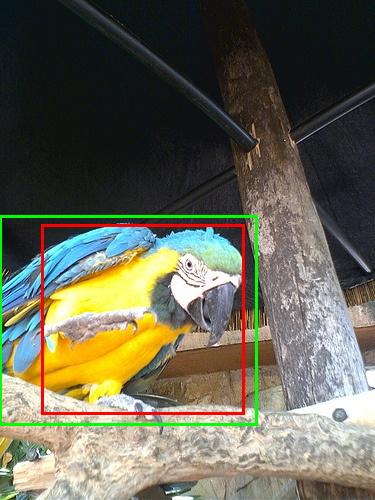}\\
	\medskip
	\includegraphics[height=3.2cm]{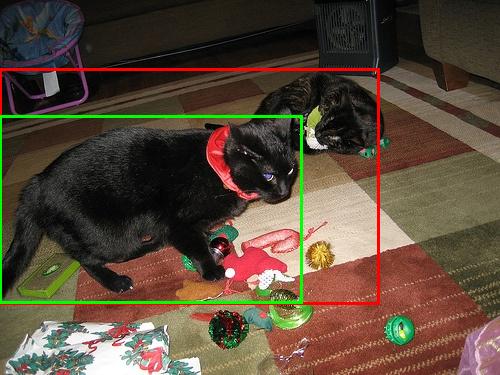}
	\hfill
	\includegraphics[height=3.2cm]{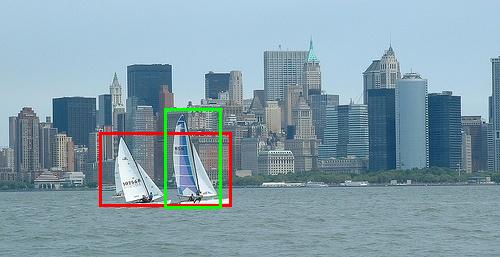}
	\hfill
	\includegraphics[height=3.2cm]{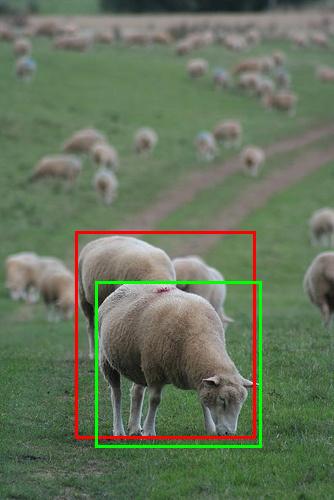}
	\caption{\textbf{Example detections of R-FCN (red) and DP-FCN (green).} DP-FCN tightly fits objects (first row) and separates close instances (second row) better than R-FCN.}
	\label{fig:examples}
\end{figure}

Detailed breakdowns of false positives are provided in Fig.~\ref{fig:pr} for R-FCN and DP-FCN.\footnote{See \url{http://mscoco.org/dataset/\#detections-eval} for full details of metrics.}
We see that the biggest gain comes from reduced localization errors (C75 and C50 metrics), and the corresponding curves are higher for DP-FCN. Ignoring those errors, recognition accuracy is consistently around 1 point better (Loc and Oth metrics). However, both models roughly keep the same number of false negatives (BG metric).

Examples of detection outputs are illustrated in Fig.~\ref{fig:examples} to visually evaluate proposed improvements. It appears that R-FCN can more easily miss extremal parts of objects (see first row, \eg the right wing of the plane) and that DP-FCN is better at separating close instances (see second row, \eg the two sheep one behind the other), thanks to deformable parts.

\vspace{-1em}
\paragraph{Interpretation of parts.}

\begin{figure}[t]
	\centering
	\includegraphics[height=4cm]{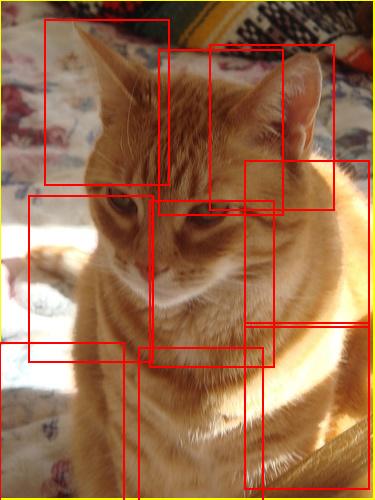}
	\hfill
	\includegraphics[height=4cm]{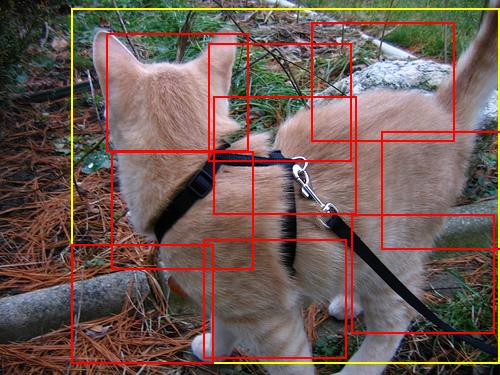}
	\hfill
	\includegraphics[height=4cm]{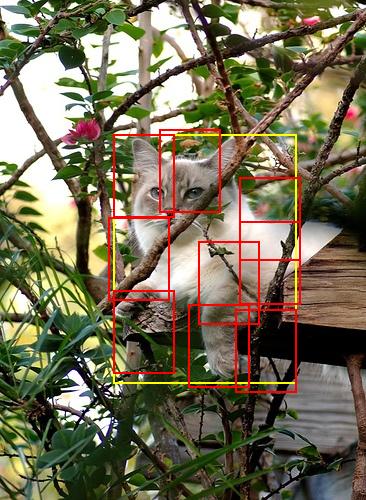}
	\caption{\textbf{Examples of deformations of parts.} Initial region proposals are shown in yellow and deformed parts in red. Only $3\times3$ parts are displayed for clarity.}
	\label{fig:parts}
\end{figure}

As in the original DPM~\cite{felzenszwalb2010object}, the semantics of parts is not explicit in our model. Part positions are instead automatically learned to optimize detection performances, in a weakly supervised manner. Therefore the interpretation in terms of semantic parts is not systematic, especially because our division of regions into parts is finer than in DPM, leading to smaller part areas. Some deformed parts are displayed on Fig.~\ref{fig:parts} with a $3\times3$ part division for easier visualization. It is noticeable that DP-FCN is able to better fit to objects with deformable parts than with simple bounding boxes.

\vspace{-1em}
\paragraph{Network architecture.}

We compare DP-FCN with several FCN backbone architectures in Tab.~\ref{tab:net}, in particular the 50- and 101-layer versions of ResNet~\cite{he2016deep}, Wide ResNet~\cite{zagoruyko2016wide} and ResNeXt~\cite{xie2017aggregated}.
We see that the detection mAP of DP-FCN can be significantly increased by using better networks.
ResNeXt-101 (64x4d) gives the best results among the tested ones, with large improvements in all metrics, despite not using dilated convolutions.

\begin{table}[h]
	\begin{center}
		\begin{tabular}{@{}l|ccc@{}}
			\toprule
			FCN architecture for DP-FCN & mAP@0.5 & mAP@0.75 & mAP@[0.5:0.95] \\
			\midrule
			ResNet-50~\cite{he2016deep} & 76.1 & 40.9 & 41.3 \\
			ResNeXt-50 (32x4d)~\cite{xie2017aggregated}$^\star$ & 76.3 & 40.8 & 41.4 \\
			Wide ResNet-50-2~\cite{zagoruyko2016wide} & 77.9 & 43.3 & 42.9 \\
			ResNet-101~\cite{he2016deep} & 78.1 & 44.2 & 43.6 \\
			ResNeXt-101 (32x4d)~\cite{xie2017aggregated}$^\star$ & 78.6 & 45.2 & 44.4 \\
			ResNeXt-101 (64x4d)~\cite{xie2017aggregated}$^\star$ & \textbf{79.5} & \textbf{47.8} & \textbf{45.7} \\
			\bottomrule
		\end{tabular}
	\end{center}
	\caption{\textbf{Comparison of DP-FCN with different FCN architectures} on PASCAL VOC 2007 test in average precision (\%). Entries marked with $^\star$ do not use dilated convolutions.}
	\label{tab:net}
\end{table}

\vspace{-1em}
\subsection{PASCAL VOC results}
\label{sec:xp}

\paragraph{Experimental setup.}

We bring the following improvements to the setup of Section~\ref{sec:ablation}, the details of which are in Appendix~\ref{sec:xpsetup_voc}: we use ResNeXt-101 (64x4d)~\cite{xie2017aggregated} and increase the number of iterations.
We include common tricks: color data augmentations~\cite{krizhevsky2012imagenet}, bounding box voting~\cite{gidaris2015object}, and averaging of detections between original and flipped images~\cite{bell2016inside, zagoruyko2016a}. We set the relative weight of the multi-task (classification/localization) loss~\cite{girshick2015fast} to 7 and enlarge input boxes by a factor 1.3 to include some context.

\vspace{-1em}
\paragraph{PASCAL VOC 2007 and 2012.}

Results of DP-FCN along with those of recent methods are reported in Tab.~\ref{tab:res_voc07} for VOC 2007 and in Tab.~\ref{tab:res_voc12} for VOC 2012. For fair comparisons we only report results of methods trained on VOC07+12 and VOC07++12 respectively, but using additional data, \eg COCO images, usually improves results~\cite{he2016deep, dai2016r}. DP-FCN achieves 83.1\% and 80.9\% on these two datasets, yielding large gaps with all competing methods. In particular, DP-FCN outperforms R-FCN~\cite{dai2016r}, the work closest to ours, by significant margins (2.6\% and 3.3\% respectively). We note that these results could be further improved with additional common enhancements, \eg multi-scale training and testing~\cite{he2015spatial} or OHEM~\cite{shrivastava2016training}.

\begin{table}[t]
	\begin{center}
	\begin{adjustbox}{max width=\textwidth}
	\begin{tabular}{@{}l|c|cccccccccccccccccccc@{}}
		\toprule
		Method & mAP & {\footnotesize aero} & {\footnotesize bike} & {\footnotesize bird} & {\footnotesize boat} & {\footnotesize bottle} & {\footnotesize bus} & {\footnotesize car} & {\footnotesize cat} & {\footnotesize chair} & {\footnotesize cow} & {\footnotesize table} & {\footnotesize dog} & {\footnotesize horse} & {\footnotesize mbike} & {\footnotesize person} & {\footnotesize plant} & {\footnotesize sheep} & {\footnotesize sofa} & {\footnotesize train} & {\footnotesize tv} \\
		\midrule
		FRCN~\cite{girshick2015fast} & 70.0 & 77.0 & 78.1 & 69.3 & 59.4 & 38.3 & 81.6 & 78.6 & 86.7 & 42.8 & 78.8 & 68.9 & 84.7 & 82.0 & 76.6 & 69.9 & 31.8 & 70.1 & 74.8 & 80.4 & 70.4 \\
		HyperNet~\cite{kong2016hypernet} & 76.3 & 77.4 & 83.3 & 75.0 & 69.1 & 62.4 & 83.1 & 87.4 & 87.4 & 57.1 & 79.8 & 71.4 & 85.1 & 85.1 & 80.0 & 79.1 & 51.2 & 79.1 & 75.7 & 80.9 & 76.5 \\
		Faster R-CNN~\cite{ren2015faster} & 76.4 & 79.8 & 80.7 & 76.2 & 68.3 & 55.9 & 85.1 & 85.3 & 89.8 & 56.7 & 87.8 & 69.4 & 88.3 & 88.9 & 80.9 & 78.4 & 41.7 & 78.6 & 79.8 & 85.3 & 72.0 \\
		SSD~\cite{liu2016ssd} & 76.8 & 82.4 & 84.7 & 78.4 & 73.8 & 53.2 & 86.2 & 87.5 & 86.0 & 57.8 & 83.1 & 70.2 & 84.9 & 85.2 & 83.9 & 79.7 & 50.3 & 77.9 & 73.9 & 82.5 & 75.3 \\
		MR-CNN~\cite{gidaris2015object} & 78.2 & 80.3 & 84.1 & 78.5 & 70.8 & 68.5 & 88.0 & 85.9 & 87.8 & 60.3 & 85.2 & 73.7 & 87.2 & 86.5 & 85.0 & 76.4 & 48.5 & 76.3 & 75.5 & 85.0 & 81.0 \\
		LocNet~\cite{gidaris2016locnet} & 78.4 & 80.4 & 85.5 & 77.6 & 72.9 & 62.2 & 86.8 & 87.5 & 88.6 & 61.3 & 86.0 & 73.9 & 86.1 & 87.0 & 82.6 & 79.1 & 51.7 & 79.4 & 75.2 & 86.6 & 77.7 \\
		FRCN OHEM~\cite{shrivastava2016training} & 78.9 & 80.6 & 85.7 & 79.8 & 69.9 & 60.8 & 88.3 & 87.9 & 89.6 & 59.7 & 85.1 & \textbf{76.5} & 87.1 & 87.3 & 82.4 & 78.8 & 53.7 & 80.5 & 78.7 & 84.5  & 80.7 \\
		ION~\cite{bell2016inside} & 79.4 & 82.5 & 86.2 & 79.9 & 71.3 & 67.2 & 88.6 & 87.5 & 88.7 & 60.8 & 84.7 & 72.3 & 87.6 & 87.7 & 83.6 & 82.1 & 53.8 & 81.9 & 74.9 & 85.8 & 81.2 \\
		R-FCN~\cite{dai2016r} & 80.5 & 79.9 & 87.2 & 81.5 & 72.0 & 69.8 & 86.8 & 88.5 & 89.8 & \textbf{67.0} & \textbf{88.1} & 74.5 & 89.8 & 90.6 & 79.9 & 81.2 & 53.7 & 81.8 & \textbf{81.5} & 85.9 & 79.9 \\

		DP-FCN~[ours] & \textbf{83.1} & \textbf{89.8} & \textbf{88.6} & \textbf{85.2} & \textbf{73.9} & \textbf{74.7} & \textbf{92.1} & \textbf{90.4} & \textbf{94.4} & 58.3 & 84.9 & 75.2 & \textbf{93.4} & \textbf{93.1} & \textbf{87.4} & \textbf{85.9} & \textbf{53.9} & \textbf{85.3} & 80.0 & \textbf{90.4} & \textbf{85.9} \\
		\bottomrule
	\end{tabular}
	\end{adjustbox}
	\end{center}
	\caption{\textbf{Detailed detection results on PASCAL VOC 2007 test} in average precision (\%). For fair comparisons, the table only includes methods trained on PASCAL VOC 07+12.}
	\label{tab:res_voc07}
\end{table}

\begin{table}[t]
	\begin{center}
	\begin{adjustbox}{max width=\textwidth}
	\begin{tabular}{@{}l|c|cccccccccccccccccccc@{}}
		\toprule
		Method & mAP & {\footnotesize aero} & {\footnotesize bike} & {\footnotesize bird} & {\footnotesize boat} & {\footnotesize bottle} & {\footnotesize bus} & {\footnotesize car} & {\footnotesize cat} & {\footnotesize chair} & {\footnotesize cow} & {\footnotesize table} & {\footnotesize dog} & {\footnotesize horse} & {\footnotesize mbike} & {\footnotesize person} & {\footnotesize plant} & {\footnotesize sheep} & {\footnotesize sofa} & {\footnotesize train} & {\footnotesize tv} \\
		\midrule
		FRCN~\cite{girshick2015fast} & 68.4 & 82.3 & 78.4 & 70.8 & 52.3 & 38.7 & 77.8 & 71.6 & 89.3 & 44.2 & 73.0 & 55.0 & 87.5 & 80.5 & 80.8 & 72.0 & 35.1 & 68.3 & 65.7 & 80.4 & 64.2 \\
		HyperNet~\cite{kong2016hypernet} & 71.4 & 84.2 & 78.5 & 73.6 & 55.6 & 53.7 & 78.7 & 79.8 & 87.7 & 49.6 & 74.9 & 52.1 & 86.0 & 81.7 & 83.3 & 81.8 & 48.6 & 73.5 & 59.4 & 79.9 & 65.7 \\
		Faster R-CNN~\cite{ren2015faster} & 73.8 & 86.5 & 81.6 & 77.2 & 58.0 & 51.0 & 78.6 & 76.6 & 93.2 & 48.6 & 80.4 & 59.0 & 92.1 & 85.3 & 84.8 & 80.7 & 48.1 & 77.3 & 66.5 & 84.7 & 65.6 \\
		SSD~\cite{liu2016ssd} & 74.9 & 87.4 & 82.3 & 75.8 & 59.0 & 52.6 & 81.7 & 81.5 & 90.0 & 55.4 & 79.0 & 59.8 & 88.4 & 84.3 & 84.7 & 83.3 & 50.2 & 78.0 & 66.3 & 86.3 & 72.0 \\
		FRCN OHEM~\cite{shrivastava2016training} & 76.3 & 86.3 & \textbf{85.0} & 77.0 & 60.9 & 59.3 & 81.9 & 81.1 & 91.9 & 55.8 & 80.6 & \textbf{63.0} & 90.8 & 85.1 & 85.3 & 80.7 & 54.9 & 78.3 & 70.8 & 82.8 & 74.9 \\
		ION~\cite{bell2016inside} & 76.4 & 88.0 & 84.6 & 77.7 & 63.7 & 63.6 & 80.8 & 80.8 & 90.9 & 55.5 & 81.9 & 60.9 & 89.1 & 84.9 & 84.2 & 83.9 & 53.2 & 79.8 & 67.4 & 84.4 & 72.9 \\
		R-FCN~\cite{dai2016r} & 77.6 & 86.9 & 83.4 & 81.5 & 63.8 & 62.4 & 81.6 & 81.1 & 93.1 & 58.0 & 83.8 & 60.8 & \textbf{92.7} & 86.0 & 84.6 & 84.4 & 59.0 & 80.8 & 68.6 & 86.1 & 72.9 \\
		DP-FCN~[ours]\footnotemark & \textbf{80.9} & \textbf{89.3} & 84.2 & \textbf{85.4} & \textbf{74.4} & \textbf{70.0} & \textbf{84.0} & \textbf{86.2} & \textbf{93.9} & \textbf{62.9} & \textbf{85.1} & 62.7 & \textbf{92.7} & \textbf{87.4} & \textbf{86.0} & \textbf{86.8} & \textbf{61.3} & \textbf{85.1} & \textbf{74.8} & \textbf{88.2} & \textbf{78.5} \\
		\bottomrule
	\end{tabular}
	\end{adjustbox}
	\end{center}
	\caption{\textbf{Detailed detection results on PASCAL VOC 2012 test} in average precision (\%). For fair comparisons, the table only includes methods trained on PASCAL VOC 07++12.}
	\label{tab:res_voc12}
\end{table}

\section{Conclusion}

In this paper, we propose DP-FCN, a new deep model for object detection. While traditional region-based detectors use generic bounding boxes to extract features from, DP-FCN is more flexible and focuses on discriminative elements to align them. It learns a part-based representation of objects in an efficient way with a natural integration into FCNs and without any additional annotations during training. This improves both recognition by building invariance to local transformations, and localization thanks to a dedicated module explicitly leveraging computed positions of parts to refine predictions with geometric information. Experimental validation shows significant gains on several common metrics. As a future work, we will test our model on a larger-scale dataset, such as MS COCO~\cite{lin2014microsoft}.\footnotetext{\url{http://host.robots.ox.ac.uk:8080/anonymous/QNUYVS.html}}

\appendix

\section{Implementation details}

\subsection{Deformable part-based RoI pooling layer}
\label{sec:impl_dproi}

We normalize the displacements $(dx,dy)$ by the widths and heights of parts to make the layer invariant to the scales of the images. We also normalize the classification feature maps before forwarding them to deformable part-based RoI pooling layer to ensure classification and regularization terms are comparable. We do this by $L_2$-normalizing at each spatial location the block of $C+1$ maps for each part separately, \ie replacing $z$ from Eq.~\eqref{eq:dproi} with
\begin{equation}
	\bar{z}_{i,j,c}(x,y) = \frac{z_{i,j,c}(x,y)}{\sqrt{\sum_{c'} z_{i,j,c'}(x,y)^{2}}}.
	\label{eq:norm}
\end{equation}
Optimization of $(dx,dy)$ is performed by brute force in limited ranges and not whole images. With $\lambda^{def}$ (Eq.~\eqref{eq:dproi}) not too small, the regularization effectively restricts values of the displacements, leaving the results of pooling unchanged.
In all experiments, we use $\lambda^{def} = 0.3$.

\subsection{Deformation-aware localization refinement}
\label{sec:impl_loc}

The localization module is applied for each class separately and takes the normalized displacements $d_c^{R}$ of a class as input, of size $2k^{2}$ (\ie a 2D displacement for each part). It is composed of two fully connected layers with a ReLU between them. The size of the first layer is set to 256 in all our experiments.
The output from average pooling (upper path in Fig.~\ref{fig:loc_ref}) is the main outcome and is obtained from the visual features only without considering deformations. The one from the fully connected layers (lower path in Fig.~\ref{fig:loc_ref}) encodes the positions of parts, and is merged with the first with an element-wise product (both are of size 4 for each class) to adjust it accordingly to the exact locations where it was computed.

\section{Experimental setups}

\subsection{Ablation study}
\label{sec:xpsetup_ab}

We use the fully convolutional backbone architecture ResNet-50~\cite{he2016deep} whose model pre-trained on ImageNet is freely available. The network is trained with SGD for 60,000 iterations with a learning rate of $5\cdot10^{-4}$ and for 20,000 further iterations with $5\cdot10^{-5}$. The momentum parameter is set to 0.9 and the weight decay to $10^{-4}$. Each mini-batch is composed of 64 regions from a single image at the scale of 600 px, selected according to Fast R-CNN~\cite{girshick2015fast}. Horizontal flipping of images with probability 0.5 is used as data augmentation. We exploit the region proposals computed by AttractioNet~\cite{gidaris2016locnet, gidaris2016attend}, released by the authors. The top 2,000 regions are used for learning and the top 300 are evaluated during inference. We use $k \times k = 7 \times 7$ parts, as advised by the authors of R-FCN~\cite{dai2016r}. As is common practice, detections are post-processed with NMS.

\subsection{PASCAL VOC results}
\label{sec:xpsetup_voc}

Changes with respect to the previous setup include replacing ResNet-50 by ResNeXt-101 (64x4d)~\cite{xie2017aggregated}, increasing the number of iterations to 120,000 and 160,000 with the same learning rates, using 2 images per mini-batch with the same number of regions per image.
We also include common tricks as described in the main paper.

\section{Examples of detections with DP-FCN}

Below are some example detections (using VOC color code) on unseen VOC 2007 test images, from the final DP-FCN model trained on VOC 07+12 data (Section~\ref{sec:xp}).

\begin{figure}[H]
	\centering
	\includegraphics[height=3cm]{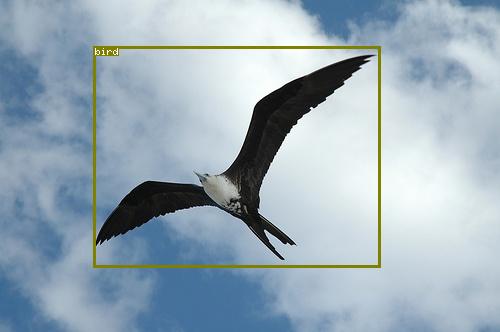}
	\hfill
	\includegraphics[height=3cm]{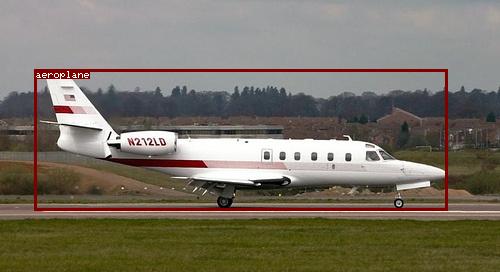}
	\hfill
	\includegraphics[height=3cm]{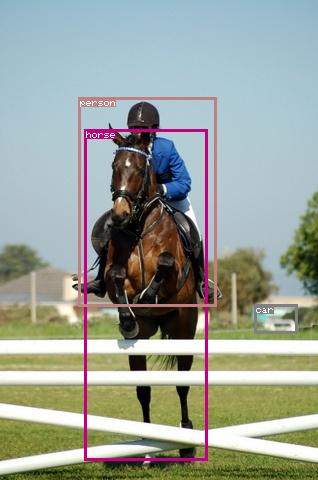}\\
	\medskip
	\includegraphics[height=2.8cm]{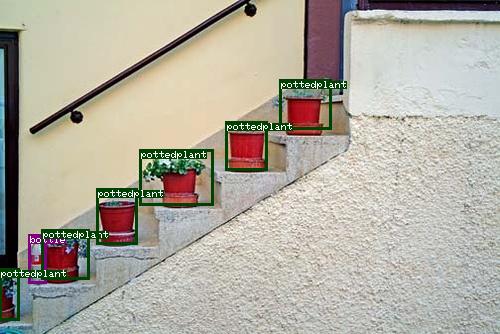}
	\hfill
	\includegraphics[height=2.8cm]{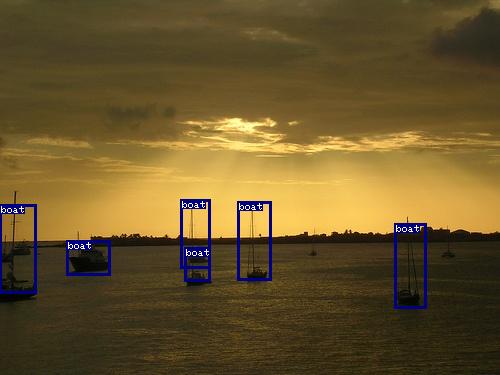}
	\hfill
	\includegraphics[height=2.8cm]{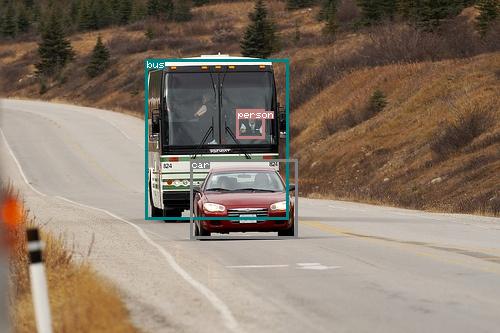}\\
	\medskip
	\includegraphics[height=2.8cm]{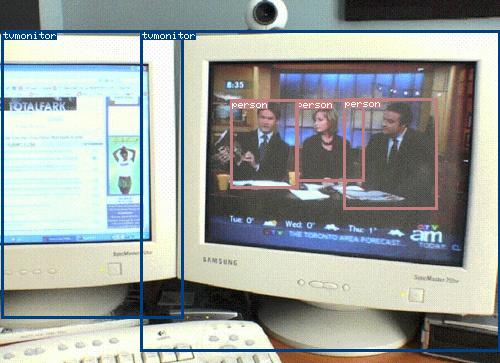}
	\hfill
	\includegraphics[height=2.8cm]{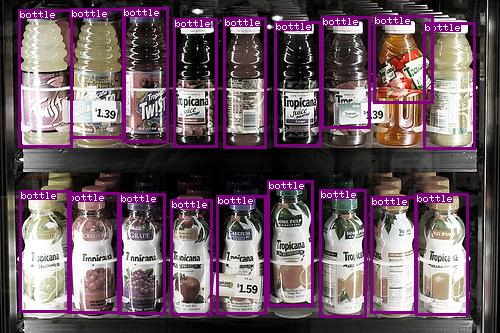}
	\hfill
	\includegraphics[height=2.8cm]{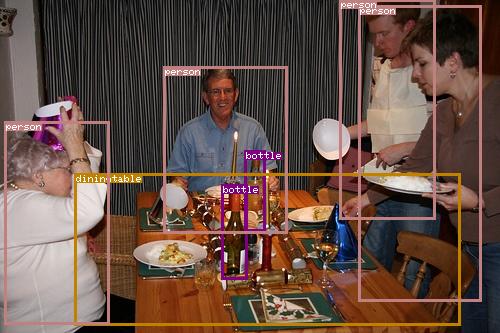}\\
	\medskip
	\includegraphics[height=3.4cm]{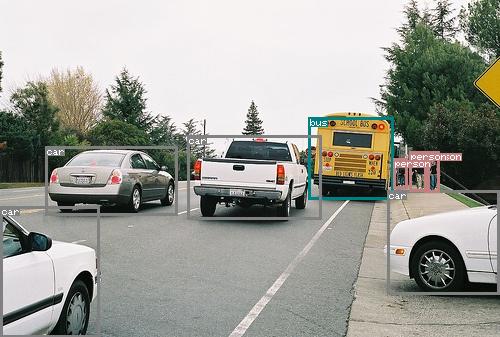}
	\hfill
	\includegraphics[height=3.4cm]{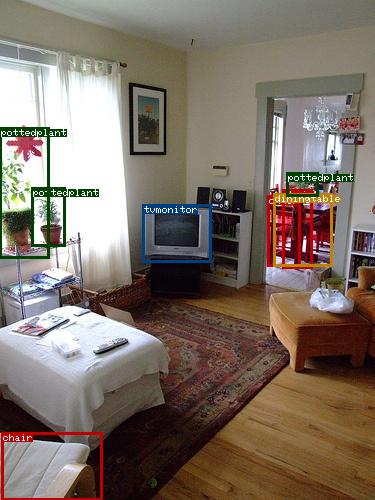}
	\hfill
	\includegraphics[height=3.4cm]{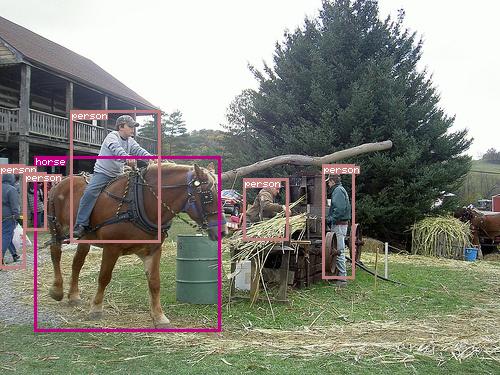}\\
	\medskip
	\includegraphics[height=2.9cm]{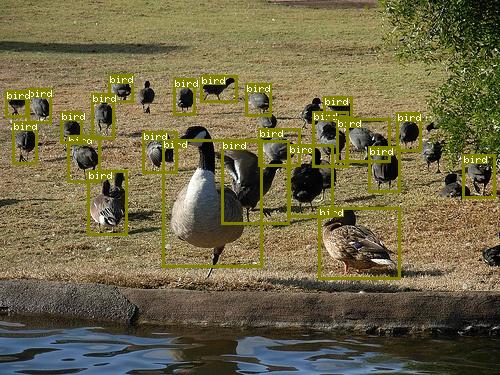}
	\hfill
	\includegraphics[height=2.9cm]{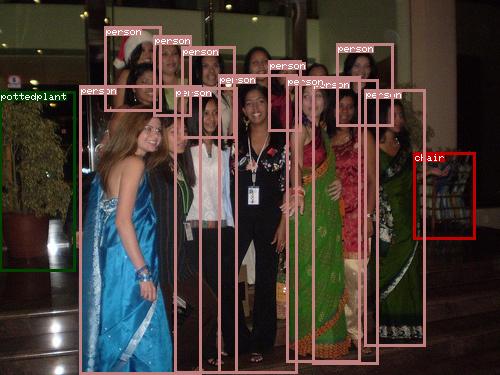}
	\hfill
	\includegraphics[height=2.9cm]{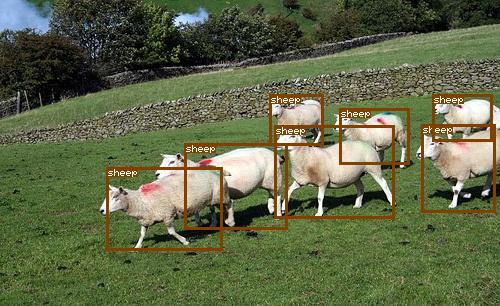}
\end{figure}

\vspace*{\fill}
\begin{figure}[!htb]
	\includegraphics[height=3.1cm]{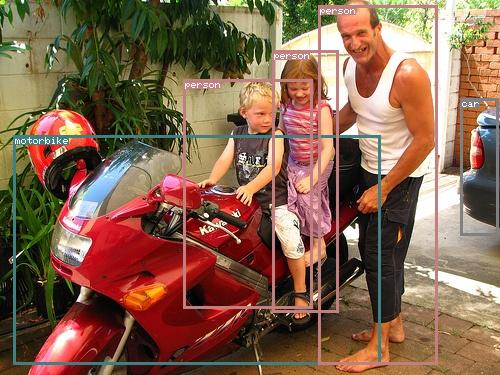}
	\hfill
	\includegraphics[height=3.1cm]{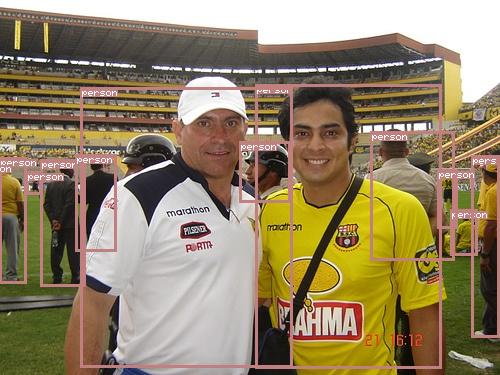}
	\hfill
	\includegraphics[height=3.1cm]{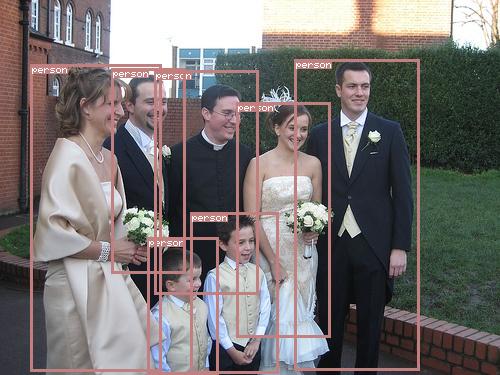}\\
	\medskip
	\includegraphics[height=3.2cm]{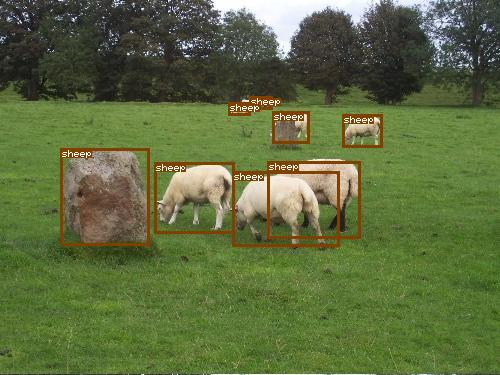}
	\hfill
	\includegraphics[height=3.2cm]{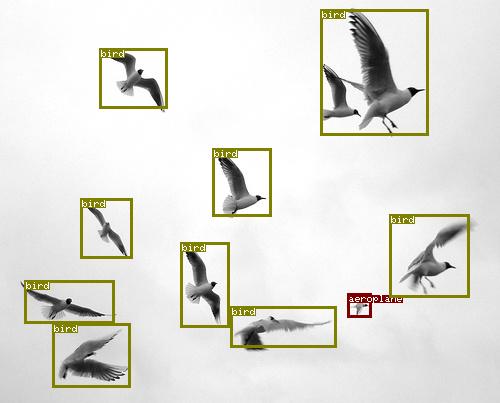}
	\hfill
	\includegraphics[height=3.2cm]{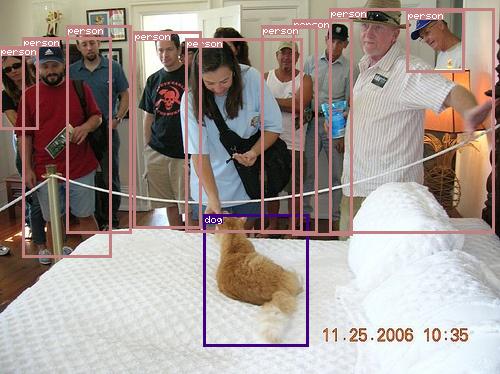}
\end{figure}
\vspace*{\fill}

\bibliography{./biblio}
\end{document}